\documentclass[letterpaper, 10 pt, journal, twoside]{IEEEtran}

\usepackage{threeparttable}  
\usepackage{graphics} 
\usepackage{graphicx}
\usepackage{xcolor}
\usepackage{svg}
\usepackage{epsfig} 
\usepackage{amsmath} 
\usepackage{amssymb}  
\usepackage{bm}
\usepackage{mathtools}
\usepackage{ragged2e}
\usepackage{float}
\usepackage{algorithm}
\usepackage[noend]{algpseudocode}
\newcommand\Algphase[1]{%
    \vspace*{-0.7\baselineskip}%
    \Statex\hspace*{-\algorithmicindent}\rule{\dimexpr\linewidth+\algorithmicindent}{0.4pt}%
    \Statex\hspace*{-\algorithmicindent}\textbf{#1}%
    \vspace*{-0.7\baselineskip}%
    \Statex\hspace*{-\algorithmicindent}\rule{\dimexpr\linewidth+\algorithmicindent}{0.4pt}%
}

\usepackage{color}
\usepackage{citesort}
\usepackage{cite}
\usepackage{flushend}
\usepackage{url}
\usepackage[breaklinks]{hyperref}
\usepackage[capitalise]{cleveref}
\usepackage{orcidlink}
\usepackage{booktabs}
\usepackage{makecell}
\usepackage{multirow}
\usepackage{lettrine}
\usepackage{siunitx}
\usepackage[nolist,nohyperlinks]{acronym}
\acrodef{method}[AOM]{ACRONYM OF METHOD}
\acrodef{gnss}[GNSS]{Global Navigation Satellite System}
\acrodef{ransac}[RANSAC]{Random Sample Consensus}
\acrodef{slam}[SLAM]{Simultaneous Localization And Mapping}
\acrodef{pca}[PCA]{Principal Component Analysis}
\acrodef{ekf}[EKF]{Extended Kalman Filter}
\acrodef{rmse}[RMSE]{Root Mean Square Error} 
\acrodef{ape}[APE]{Absolute Pose Error}
\acrodef{cfar}[CFAR]{Constant False Alarm Rate}
\acrodef{snr}[SNR]{Signal to Noise Ratio}
\acrodef{rcs}[RCS]{Radar Cross Section}
\acrodef{imu}[IMU]{Inertial Measurement Unit}
\acrodef{sgm}[SGM]{Segmi-Global Matching}
\acrodef{dnn}[DNN]{Deep Neural Network}
\acrodef{gru}[GRU]{Gated Recurrent Unit}
\acrodef{hpr}[HPR]{Hidden Point Removal}
\acrodef{raft}[RAFT]{Recurrent All-Pairs Field Transforms}
\acrodef{fov}[FOV]{Field of View}
\acrodef{mclab}[MC-lab]{Marine Cybernetics laboratory}
\acrodef{vio}[VIO]{Visual-Inertial Odometry}
\acrodef{rcm}[RCM]{Refractive Camera Model}
\acrodef{sfm}[SFM]{Structure from Motion}
\acrodef{vae}[VAE]{Variational Autoencoder}
\acrodef{mse}[MSE]{Mean Squared Error}

\newcommand{\Eq}{Eq.}

\usepackage{tikz}
\usetikzlibrary{positioning, shapes.geometric, arrows.meta, decorations.pathreplacing, calligraphy}
\DeclareMathAlphabet{\pazocal}{OMS}{zplm}{m}{n}

\DeclareMathAlphabet{\mathpzc}{OT1}{pzc}{m}{it}

\usepackage{array}
\newcolumntype{C}[1]{>{\centering\arraybackslash}p{#1}}
\newcolumntype{M}[1]{>{\raggedright\arraybackslash}p{#1}}

\usepackage{array} 
\newcolumntype{L}[1]{>{\raggedright\let\newline\\\arraybackslash\hspace{0pt}}m{#1}}	
\newcolumntype{S}[1]{>{\centering\let\newline\\\arraybackslash\hspace{0pt}}m{#1}}
\newcolumntype{R}[1]{>{\raggedleft\let\newline\\\arraybackslash\hspace{0pt}}m{#1}}

\makeatletter
\renewcommand*{\@opargbegintheorem}[3]{\trivlist
  \item[\hskip \labelsep{\itshape #1\ #2}] \textit{(#3)}\ }
\makeatother

\newcommand{\addition}[1]{\textcolor{black}{#1}}

\begin{document}

    \title{Collaborative Exploration with a Marsupial Ground-Aerial Robot Team through Task-Driven Map Compression}

    \author{Angelos Zacharia, Mihir Dharmadhikari, and Kostas Alexis%
    \thanks{Manuscript received: March, 21, 2025; Revised June, 18, 2025; Accepted September, 02, 2025.}%
    \thanks{This paper was recommended for publication by Editor Olivier Stasse upon evaluation of the Associate Editor and Reviewers' comments. This work was supported by the European Commission Horizon Europe grants SYNERGISE (EC 101121321), SPEAR (EC 101119774) and DIGIFOREST (EC 101070405).}%
    \thanks{The authors are with the Autonomous Robots Lab, Norwegian University of Science and Technology (NTNU), Norway {\tt\footnotesize angelos.zacharia@ntnu.no}}%
    \thanks{Digital Object Identifier (DOI): see top of this page.}}

    \markboth{IEEE Robotics and Automation Letters. Preprint Version. Accepted September, 2025}
    {Zacharia \MakeLowercase{\textit{et al.}}: Collaborative Exploration with a Marsupial Ground-Aerial Robot Team through Task-Driven Map Compression} 
    
    \maketitle

    \begin{abstract}
        Efficient exploration of unknown environments is crucial for autonomous robots, especially in confined and large-scale scenarios with limited communication. To address this challenge, we propose a collaborative exploration framework for a marsupial ground-aerial robot team that leverages the complementary capabilities of both platforms. The framework employs a graph-based path planning algorithm to guide exploration and deploy the aerial robot in areas where its expected gain significantly exceeds that of the ground robot, such as large open spaces or regions inaccessible to the ground platform, thereby maximizing coverage and efficiency. To facilitate large-scale spatial information sharing, we introduce a bandwidth-efficient, task-driven map compression strategy. This method enables each robot to reconstruct resolution-specific volumetric maps while preserving exploration-critical details, even at high compression rates. By selectively compressing and sharing key data, communication overhead is minimized, ensuring effective map integration for collaborative path planning. Simulation and real-world experiments validate the proposed approach, demonstrating its effectiveness in improving exploration efficiency while significantly reducing data transmission.
    \end{abstract}

    \IEEEpeerreviewmaketitle
    \section{Introduction}

    \IEEEPARstart{A}{dvancements} in robotic systems have facilitated their deployment across a wide range of autonomous missions. Both aerial and ground robots are now extensively utilized for various applications, including search and rescue~\cite{delmerico2019current}, surveillance~\cite{grocholsky2006cooperative}, inspection~\cite{dharmadhikari2023semantics}, and exploration~\cite{kulkarni2022autonomous}.

    Efficiently exploring unknown environments remains a significant challenge, whether navigating confined indoor spaces or traversing expansive outdoor landscapes. Single-robot systems often encounter limitations in terms of speed, sensing range, and their ability to navigate complex terrains. These challenges can be effectively addressed through the deployment of heterogeneous robot teams, where the strengths of one robot complement the limitations of the others. Marsupial ground-aerial systems exemplify a collaborative approach, combining the robustness and load capacity of ground robots with the agility and three-dimensional mobility of aerial robots. This synergy enables the strategic deployment of the aerial robot to efficiently explore large open spaces and high-ceiling environments, providing rapid coverage and accessing areas beyond the ground robot’s reach for seamless mapping. 

    \begin{figure}[tp]
        \centering
        \includegraphics[clip, trim=0cm 0cm 0cm 0cm, width=1\linewidth]{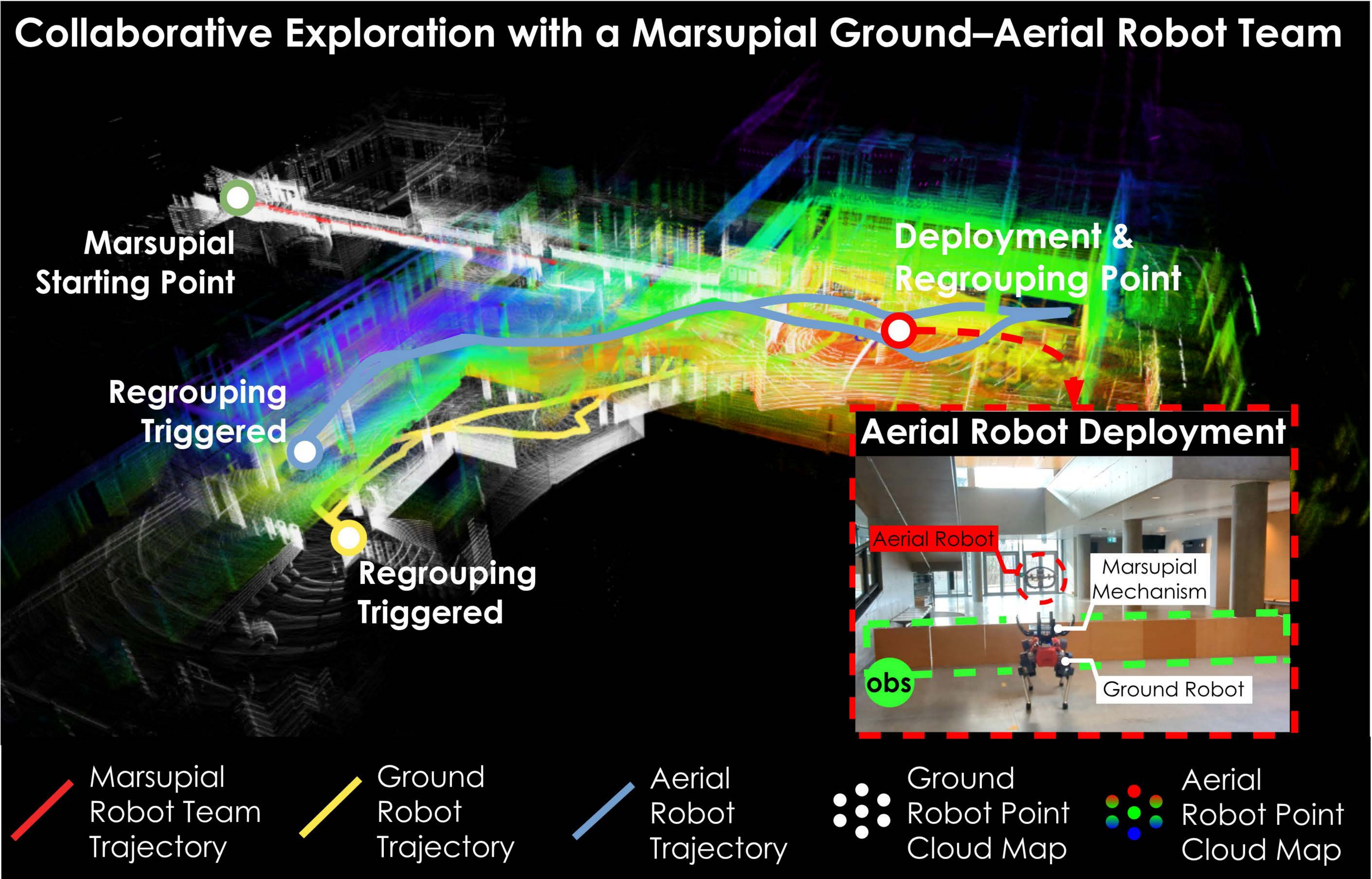}
        \vspace{-2em}
        \caption{\addition{Combined map from a real-world experiment, where the marsupial ground–aerial robot team collaborated in exploration. The aerial robot took over after a physical obstacle (obs) prevented the ground robot's progress.}}
        \label{fig:intro}
        \vspace{-2ex}
    \end{figure}

    \addition{Collaborative exploration relies on efficient data sharing between robots. Real-time exchange of sensor data, maps, and positions enables coordination and prevents redundant work. However, differences in processing power, sensor types, and communication constraints—such as limited bandwidth, high latency, and intermittent links—pose significant challenges, especially in complex environments.}

    \addition{To address these challenges, this paper presents a comprehensive framework that integrates planning and deployment strategies for a marsupial ground-aerial robot team. Unlike prior work that either focuses on communication-efficient compression of generic point cloud data or on heterogeneous robot coordination without scalable data sharing, our contributions are twofold. First, we propose a bandwidth-efficient, task-driven point cloud compression method tailored for volumetric map reconstruction at mission-relevant resolutions. By emphasizing occupancy-relevant structure over raw point cloud fidelity, our approach achieves high compression rates while retaining the information essential for planning. This method is open-sourced at \href{https://github.com/ntnu-arl/pcl-vae/}{https://github.com/ntnu-arl/pcl-vae/}. Second, we introduce a decentralized collaborative exploration framework that leverages the complementary capabilities of a marsupial robot team. It features an aerial robot deployment strategy, keyframe-based map sharing for coordinated planning, and an energy-aware regrouping strategy. The framework is validated in large-scale simulations and real-world experiments, demonstrating improvements in both exploration efficiency and bandwidth usage. More experimental results from the real-world trials can be found at \href{https://ntnu-arl.github.io/marsupial-collaborative-exploration/}{\mbox{https://ntnu-arl.github.io/marsupial-collaborative-exploration/}}}

    The remainder of this paper is organized as follows. Section \ref{sec:related_work} reviews related work on marsupial robot teams and map compression and sharing techniques. Section \ref{sec:proposed_approach} details the proposed framework, followed by evaluation studies in Section \ref{sec:evaluation_studies}. Finally, Section \ref{sec:conclusion} concludes the paper.  
    
    \section{Related Work}
\label{sec:related_work}

\subsection{Marsupial Systems-of-Systems}

    Research in marsupial robotics, particularly ground/aerial systems, has historically been sparse. However, recent advancements have emerged, notably through the efforts of several teams involved in the DARPA Subterranean Challenge~\cite{chung2023into}. The ground platforms primarily consisted of tracked vehicles~\cite{kottege2024heterogeneous}, rovers~\cite{cao2023exploring}, or legged robots~\cite{de2022marsupial}, while the aerial platforms were predominantly multirotor systems. Focusing on combined docking-and-recharging, the work in~\cite{moore2023combined} integrated a VTOL drone with a quadruped robot. Arguably the most well-known example, \addition{the Ingenuity helicopter, has completed multiple missions on Mars after being ferried and launched by the Perseverance rover~\cite{balaram2021ingenuity}.}

    Beyond system design, a set of works relates to the problem of planning for marsupial systems. These include works on path and trajectory planning of tethered aerial-ground systems~\cite{martinez2023path,capitan2024efficient}, and stochastic assignment for the deployment of multiple marsupial robots~\cite{lee2021stochastic}. The deployment of marsupial robots for multi-agent exploration is studied in~\cite{couceiro2014marsupial,de2022marsupial,best2024multi}. In this framework, the issue of communication constraints has repetitively attracted attention~\cite{couceiro2014marsupial,chung2023into}.

\subsection{Point Cloud Compression and Map Sharing}

    A set of methods have been developed to enable efficient point cloud compression, including both conventional and neural strategies. Exploiting predictive deep learning models and leveraging the image representation of LiDAR data, RIDDLE~\cite{zhou2022riddle} achieves high-degree of compression. The contribution in~\cite{tu2019point} exploits recurrent neural networks for efficient compression, while the work in~\cite{wiesmann2021deep} leverages a convolutional autoencoder learning compact feature descriptors from point clouds. A survey on deep learning-based point cloud compression is presented in~\cite{quach2022survey}. Targeting autonomous driving, \cite{sun2019novel} utilizes range image-based segmentation and clustering to reduce spatial redundancy, with video coding enhancing compression. \cite{feng2020real} exploits spatial and temporal redundancies in point clouds for real-time, high-efficiency compression.

    Beyond the general application of compression, a niche body of work exists focusing on compression for map sharing in multi-robot operations~\cite{lazaro2013multi}. RecNet~\cite{stathoulopoulos2024recnet} transforms 3D point clouds into compact range image embeddings for efficient encoding and sharing while it serves both the goal of place recognition tasks and collaborative mapping in resource-constrained settings. The work in~\cite{zheng2024real} first maps $3\textrm{D}$ point clouds into panoramas, uses event-triggered updates, and applies frequency-domain point cloud compression for efficient multi-robot systems. Departing from the current state-of-the-art, this work prioritizes a high degree of point cloud compression by encoding into a latent representation that explicitly focuses on the information necessary to reconstruct the occupancy map for planning and collision avoidance. While prior methods focus on compression ratios in the order of $10\times$ - $80\times$ \cite{stathoulopoulos2024recnet,zheng2024real,cao2025real}, our approach targets and achieves $300\times$ compression, enabling efficient map sharing for collaborative exploration in communication-constrained environments.
    
    \section{Proposed Approach}
\label{sec:proposed_approach}
    \begin{figure*}[tp]
        \centering
        \includegraphics[clip, trim=0cm 0cm 0cm 0cm, width=0.95\linewidth]{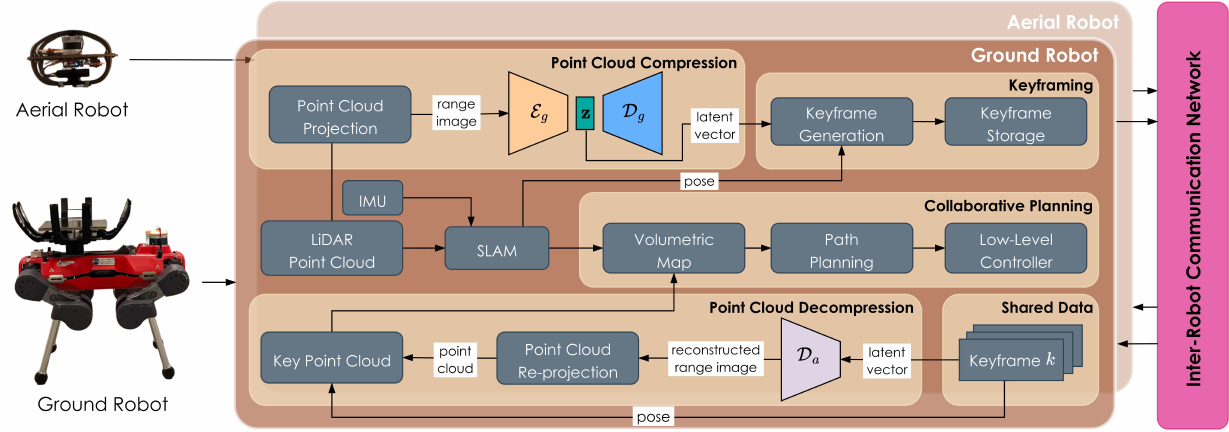}
        \vspace{-1em}
        \caption{Overview of the proposed collaborative exploration framework with bandwidth-efficient map sharing, employed by a marsupial ground-aerial robot team. As the ground robot explores while carrying the aerial robot, it continuously evaluates whether to deploy the aerial robot by comparing their respective exploration gains. At the same time, it compresses sparse point clouds, generating and storing keyframes. Upon deployment, a subset of keyframes is shared with the aerial robot to initialize its volumetric map. Both robots then explore independently, exchanging keyframes bidirectionally over the communication network. \addition{Each robot runs its own encoder to generate keyframes and uses the other’s decoder for decompression}. After a predetermined duration, both robots return to the deployment point, concluding the mission once all keyframes have been exchanged.}
        \label{fig:overview}
        \vspace{-3ex}
    \end{figure*}

    This section presents the proposed methodology utilized by a marsupial robot team to collaboratively explore unknown environments through efficient, task-driven map compression. The team consists of a ground (legged) robot and an aerial robot in a marsupial configuration, where the ground robot serves as a carrier platform for the aerial system. \addition{Both robots perform graph-based exploration path planning, with the ground robot additionally assessing the deployment of the aerial robot using an exploration gain mechanism.} Key to the proposed approach is a bandwidth-efficient map-sharing solution that enables the receiving robot to reconstruct the volumetric information acquired by the transmitting robot via the inter-robot communication network. This reconstruction allows the receiving robot to plan based on the volumetric map that integrates both its locally observed data and the shared information. To achieve this, each robot first compresses a selective subset of the point cloud data acquired by its onboard sensors, which is then transmitted along with the associated estimated pose transformations (keyframes). An overview of the proposed approach is presented in Figure~\ref{fig:overview}.
 
\subsection{Task-Driven Point Cloud Compression for Volumetric Mapping} 
    
\begin{figure*}[ht]
            \includegraphics[clip, trim=0cm 1cm 7cm 0cm, width=1.0\linewidth]{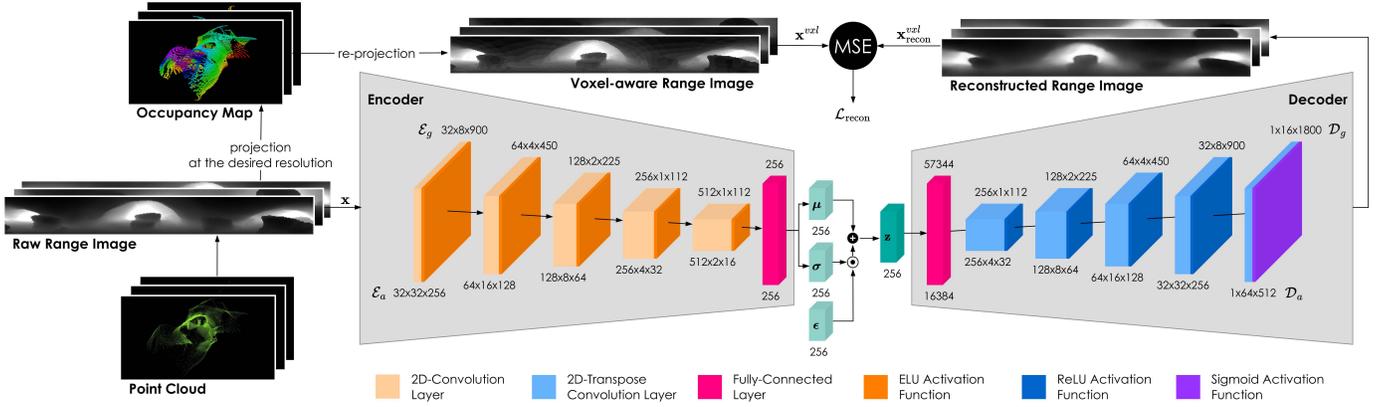}
            \vspace{-2em}
            \caption{The proposed network architecture is tailored to compressing and remapping the range image $\mathbf{x}$ to a latent vector $\mathbf{z}$, which is then used to generate the reconstructed voxel-aware range image $\mathbf{x}_{\textrm{recon}}^{vxl}$. The encoder-decoder scheme for ground robot $\{\mathcal{E}_g$,$\mathcal{D}_g\}$ and for the aerial robot $\{\mathcal{E}_a$,$\mathcal{D}_a\}$ consists of convolutional and fully connected layers along with activation functions. The output shape of each layer is indicated in the format $C\times H \times W$, representing the number of channels ($C$), height ($H$), and width ($W$), respectively.}
            \label{fig:vae_architecture}
            \vspace{-3ex}
        \end{figure*}

    \addition{Unlike conventional point cloud compression methods aimed at reconstructing raw input, we propose a task-specific solution focused on reconstructing volumetric information at a defined voxel resolution. This enables high compression rates by filtering out spatially insignificant details through voxelization, aligned with mission requirements. The proposed compression pipeline for LiDAR range data follows a two-step procedure: i) a remapping step that accounts for voxel map integration to retain only task-relevant information, and ii) a custom-trained \ac{vae} architecture that jointly remaps and encodes the input range image, along with a corresponding decoder. The proposed architecture is shown in Figure~\ref{fig:vae_architecture}.}
        
    \subsubsection{Voxel-aware Range Image Generation}

        \addition{Range images and their associated point clouds capture intricate surface geometries, especially with modern high-resolution LiDAR sensors. However, such geometric detail imposes a challenge for compression: high-frequency features can dominate the latent representation $\mathbf{z}$, consuming capacity that would otherwise encode semantically meaningful structures. To address this, we adopt a task-specific approach that preserves only the information needed to reconstruct an accurate occupancy map, rather than the full raw geometry.}
        
        \addition{We introduce a preprocessing step that converts each range image $\mathbf{x}$ in the training set $\mathbb{D}$ into a voxel-aware version $\mathbf{x}^{vxl} \in \mathbb{D}^{vxl}$, which serves as the VAE training target. Each pixel $(i,j)$ in $\mathbf{x}$ is projected into 3D space using LiDAR intrinsics and spherical-to-Cartesian conversion. The resulting 3D point cloud populates a voxelized occupancy map $\mathbf{O} \in \{0,1,2\}^{N_x \times N_y \times N_z}$ with resolution-specific voxelization $s_{vxl}$, labeling voxels as free (0), occupied (1), or unknown (2) via ray casting. Rays from the LiDAR origin mark traversed voxels as free and endpoints as occupied; untouched voxels remain unknown. To return to a 2D form, we re-trace each ray and assign the pixel in $\mathbf{x}^{vxl}$ the distance to the first occupied voxel it intersects, or mark it unknown if none is found. The resulting image appears discretized but retains meaningful structure, filtering out irrelevant fine details while preserving volumetric information critical for navigation and planning. This pipeline is efficiently implemented using NVIDIA Warp for large-scale GPU processing and performs the following operations:}
        \begin{equation}
            \addition{\forall ~\mathbf{x}\in\mathbb{D} \underset{\textrm{projection}}{\mapsto} \mathbf{O}(\mathbf{x},s_{vxl}) \underset{\textrm{re-projection}}{\mapsto} \mathbf{x}^{vxl}\in\mathbb{D}^{vxl}.}
            \label{eq:operations_for_voxelization}
        \end{equation}

    \subsubsection{Voxel-aware Range Image Compression}
        \label{sec:voxel-aware_range_image_compression}
        Motivated by the overall success of \acp{vae} and literature in task-driven compression for collision images~\cite{kulkarniTask2023}, we utilize the \ac{vae} architecture depicted in Figure~\ref{fig:vae_architecture} to learn how to simultaneously remap and compress the input raw range images such that their voxel-aware form can be reconstructed faithfully through a particularly lightweight latent space. 
        
        Let $\mathbf{x}\in\mathbb{D}$ represent a range image and $\mathbf{x}^{vxl}\in\mathbb{D}^{vxl}$ denote its corresponding voxel-aware range image, derived by applying the operations in \Eq~\eqref{eq:operations_for_voxelization}. To enable efficient dimensionality reduction of the input range image, we employ a probabilistic encoding-decoding framework that leverages the expressive power of \acp{dnn} for effective compression and simultaneously learning the voxel-aware remapping. The probabilistic decoder $p_\theta(\mathbf{x}^{vxl}|\mathbf{z})$ generates a distribution over all possible values of $\mathbf{x}^{vxl}$, given the latent representation $\mathbf{z}$ with dimensions $N_{\textbf{z}}$. Analogously, the probabilistic encoder $q_\phi(\mathbf{z}|\mathbf{x})$ learns to simultaneously encode and remap the raw range image $\mathbf{x}$ to a latent distribution with mean $\boldsymbol{\mu}\in\mathbb{R}^{N_{\textbf{z}}}$ and standard deviation $\boldsymbol{\sigma}\in(\mathbb{R}^{+})^{N_{\textbf{z}}}$. This distribution is then sampled using the reparameterization trick, $\mathbf{z} = \boldsymbol{\mu} + \boldsymbol{\sigma} \odot \boldsymbol{\epsilon}$ ($\odot$ is the element-wise multiplication operator) where $\boldsymbol{\epsilon}\sim \mathcal{N}(\mathbf{0}_{N_{\textbf{z}}},\mathbf{I}_{N_{\textbf{z}}})$ \cite{kingma2022autoencodingvariationalbayes}. Joint training of the encoder and decoder networks is guided by the loss function
        \begin{equation}
            \pazocal{L} = \pazocal{L}_{\textrm{recon}} + \beta_{\textrm{norm}}\pazocal{L}_\pazocal{\textrm{KL}}
        \end{equation}
        where $\pazocal{L}_{\textrm{recon}}$ denotes the reconstruction loss and $\pazocal{L}_{\textrm{KL}}$ represents the KL-divergence loss, both defined as
        \begin{flalign}
            &\pazocal{L}_{\textrm{recon}}(\mathbf{x}^{vxl},\mathbf{x}^{vxl}_{\textrm{recon}}) = \textrm{MSE}(\mathbf{x}^{vxl}, \mathbf{x}^{vxl}_{\textrm{recon}}) \\
            & \pazocal{L}_{\textrm{KL}}(\boldsymbol{\mu},\boldsymbol{\sigma}) = -\frac{1}{2} \sum_{n=1}^{N_{\textbf{z}}}(1+\log(\boldsymbol{\sigma}_n^2) - \boldsymbol{\mu}_n^2-\boldsymbol{\sigma}_n^2).
        \end{flalign}
        The reconstruction loss is measured as the \ac{mse} between the voxel-aware range image $\mathbf{x}^{vxl}$ and the reconstructed output $\mathbf{x}^{vxl}_{\textrm{recon}}$, excluding contributions from invalid pixels in the range image to ensure they do not affect the loss calculation. The KL-divergence loss balances the trade-off between reconstruction quality and latent space regularization by ensuring that the posterior distribution $q_\phi(\mathbf{z}|\mathbf{x})$ remains close to a predefined prior $p(\mathbf{z})$, modeled as a standard Gaussian $\mathcal{N}(\mathbf{0}_{N_{\textbf{z}}},\mathbf{I}_{N_{\textbf{z}}})$ \cite{higgins2017betavae}. The contribution of KL-divergence loss is adjusted by the tunable hyperparameter $\beta_{\textrm{norm}} = \frac{\beta\cdot N_{\textbf{z}}}{H\cdot W}$, where $\beta = 1$, and $H$ and $W$ are the height and width of the range image, respectively.
        
        In the neural architecture of the trained compression model, the encoder comprises five convolutional layers with ELU activation functions, designed to progressively reduce the spatial dimensions of $\mathbf{x}$ while increasing feature richness. At the final stage of the encoder, a fully connected layer generates two output vectors representing the mean $\boldsymbol{\mu}$ and standard deviation $\boldsymbol{\sigma}$, which parameterize the latent space distribution. The decoder adopts a symmetric structure to the encoder, beginning with a fully connected layer that transforms the latent vector $\mathbf{z}$ into an intermediate feature representation. This is followed by five deconvolutional layers with ReLU activation functions except for the final layer, which employs a sigmoid activation function. The latter ensures that the reconstructed data matches the range of the original data. This architecture (Fig.~\ref{fig:vae_architecture}) achieves $2\times$ faster inference time compared to \cite{kulkarni2023sevae}, through parameter reduction and the removal of residual layers, while maintaining effective compression and reconstruction.

\subsection{Collaborative Exploration}

    \begin{algorithm}[!htbp]
    \caption{Collaborative Exploration Path Planning}
    \label{alg:collaborative_exploration}
    \begin{algorithmic}[1]
            \vspace{0.5cm}
            \Algphase{Phase 1: Pre-Deployment}
            \Require Ground robot pose $\xi_0^g$ and its point cloud $\mathcal{P}_0^g$
            \State $\xi_k^g$ $\gets$ $\xi_0^g$
            \State \texttt{DeploymentTriggered} $\gets$ \texttt{false}
            \While{\texttt{not DeploymentTriggered}}
                \State $\xi_0^g$ $\gets$ \textbf{GetCurrentConfiguration}()
                \State $\mathbb{G}_L^g$ $\gets$ \textbf{BuildLocalGraphGroundRobot}($\xi_0^g$)
                \State $\sigma_{L,\textrm{best}}^g$, $\phi_{L,\textrm{best}}^g$ $\gets$ \textbf{GetBestLocalPathAndGain}($\mathbb{G}_L^g$) 
                \State \addition{$\bar{\mathbb{G}}_L^a$ $\gets$ \textbf{BuildVirtual3DGraph}($\xi_0^g$)}
                \State \addition{$\mathbf{p}_{\textrm{target}}^a$, $\bar{\phi}_{L,\textrm{best}}^a$ $\gets$ \textbf{GetBestVertexAndGain}($\bar{\mathbb{G}}_L^a$)} 
                \If{$\mathcal{H}(\phi_{L,\textrm{best}}^g, \bar{\phi}_{L,\textrm{best}}^a)$}
                    \State \texttt{DeploymentTriggered} $\gets$ \texttt{true}
                \EndIf
                \State $\mathbb{K}_g$ $\gets$ \textbf{Keyframing}($g$, $\xi_0^g$, $\xi_k^g$) \Comment{Algorithm \ref{alg:keyframing}}
            \EndWhile
            \vspace{0.1cm}
            \Algphase{Phase 2: Deployment}
            \Procedure{Co-Localization}{$\xi_0^g$, $\mathcal{M}$}
            \EndProcedure
            \Procedure{Map-Sharing}{$\mathbb{K}_g$, $N_k$}
                
            \EndProcedure
            \Procedure{Target-Sharing}{$\mathbf{p}_{\textrm{target}}^a$}
                
            \EndProcedure

            \vspace{0.1cm}
            \Algphase{Phase 3: Post-Deployment}
            \Require Set of robots $\mathcal{R} = \{g, a\}$
            \While{\texttt{remainingTime}  $ > t_b$}
                \ForAll{$i$ in $\mathcal{R}$}
                    \State $\xi_0^i$ $\gets$ \textbf{GetCurrentConfiguration}()
                    \State $\mathbb{G}_L^i$ $\gets$ \textbf{BuildLocalGraph}($\xi_0^i$)
                    \State $\sigma_{L,\textrm{best}}^i$ $\gets$ \textbf{GetBestLocalPath}($\mathbb{G}_L^i$) 
                    \State $\mathbb{K}_i$ $\gets$ \textbf{Keyframing}($i$, $\xi_0^i$, $\xi_k^i$) \Comment{Algorithm \ref{alg:keyframing}}
                    \If{$\exists$ neighbor robot}
                        \State \textbf{SendUnsharedKeyframes($\mathbb{K}_i$)}
                        \If{$i = g$}    \Comment{\addition{Ground Robot}}
                            \State \addition{$\mathbb{K}_a$ $\gets$\textbf{ReceiveKeyframes}()}
                            \State \addition{$\mathbb{T}_g$ $\gets$ \textbf{ComputeTimesToKeyframes}($\mathbb{K}_g$)} 
                            \State \addition{\textbf{SendTimesToKeyframes}($\mathbb{T}_g$)}
                            \State \addition{\textbf{ReceiveRegroupingPoint}()}
                        \ElsIf{$i = a$} \Comment{\addition{Aerial Robot}}
                            \State \addition{$\mathbb{K}_g$ $\gets$\textbf{ReceiveKeyframes}()}
                            \State \addition{$\mathbb{T}_g$ $\gets$ \textbf{ReceiveTimesToKeyframes}()} 
                            \State \addition{$\mathbb{T}_a$ $\gets$ \textbf{ComputeTimesToKeyframes}($\mathbb{K}_g$)}
                            \State \addition{$\mathbb{K}_{g,\text{best}}$ $\gets$ \textbf{GetRegroupingPoint}($\mathbb{T}_g$,$\mathbb{T}_a$)}
                            \State \addition{\textbf{SendRegroupingPoint}($\mathbb{K}_{g,\text{best}}$)}
                        \EndIf
                    \EndIf
                \EndFor
            \EndWhile
            \State \addition{\textbf{ReturnToRegroupingPoint}($\mathbb{G}_G^i$), $i \in \{g, a\}$}
        \end{algorithmic}
    \end{algorithm}

    \begin{algorithm}[!htbp]
        \caption{Keyframing}
        \label{alg:keyframing}
        \begin{algorithmic}[1]
            \Function{Keyframing}{i, $\xi_0^i$, $\xi_k^i$}
                \State $\Delta t$, $\Delta r$ $\gets$ \textbf{ComputePoseDifference}($\xi_0^i$, $\xi_k^i$)
                \If{$\Delta t > \tau_t$ \textbf{or} $\Delta r > \tau_r$}
                    \State $\mathcal{P}_0^i$ $\gets$ \textbf{GetCurrentPointCloud}()
                    \State $\mathbf{x}_i$ $\gets$ \textbf{PointCloudProjection}($\mathcal{P}_0^i$)
                    \State $\mathbf{z}_i$ $\gets$ $\mathcal{E}_i(\mathbf{x}_i)$
                    \State $\mathbb{K}_i$ $\gets$ $\mathbb{K}_i \cup \{\mathbf{z}_i,\xi_0^i\}$
                    \State $\xi_k^i$ $\gets$ $\xi_0^i$
                \EndIf  
                \State \textbf{return} $\mathbb{K}_i$
            \EndFunction
        \end{algorithmic}
    \end{algorithm}

    Effective exploration of complex environments requires leveraging the strengths of heterogeneous robot teams. To fully utilize the complementary capabilities of a marsupial ground-aerial robot team, we propose a collaborative exploration path planning framework, with three phases: i) pre-deployment, ii) deployment, and iii) post-deployment, as detailed in this section. Pseudocode is provided in Algorithm~\ref{alg:collaborative_exploration}.

    The proposed framework extends GBPlanner~\cite{kulkarni2022autonomous}, a graph-based exploration planner with local and global modules using Voxblox~\cite{oleynikovaVoxblox2017} for volumetric mapping. The local planner samples 3D points to build a dense graph $\mathbb{G}_L$ and computes shortest paths $\Sigma_L$ via Dijkstra's algorithm. \addition{For ground robots, samples are projected into 2.5D to respect locomotion constraints.} The best path $\sigma_{L,\textrm{best}}$ is selected based on exploration gain $\phi_{L,\textrm{best}}$. If no informative path is found, the global planner builds a sparse graph $\mathbb{G}_G$ to guide exploration and ensure return-to-home within endurance limits. Both robots independently run GBPlanner as their exploration strategy.

    \textbf{Pre-Deployment Phase:} In this phase, the ground robot carries the aerial robot and evaluates the need for deployment during each planning iteration to enhance exploration. \addition{Specifically, the ground robot constructs a 2.5D local graph $\mathbb{G}_L^g$ and identifies the optimal path $\sigma_{L,\textrm{best}}^g$ along with the corresponding exploration gain $\phi_{L,\textrm{best}}^g$ (Algo.~\ref{alg:collaborative_exploration}, lines 5–6) \cite{kulkarni2022autonomous}. In parallel, it generates a virtual 3D local graph $\bar{\mathbb{G}}_L^a$, which approximates the graph the aerial robot would construct if it were deployed. Based on the aerial robot’s sensor specifications, the ground robot selects the vertex in $\bar{\mathbb{G}}_L^a$ with the highest exploration gain $\bar{\phi}_{L,\textrm{best}}^a$, and designates it as the potential aerial target point $\mathbf{p}_{\textrm{target}}^a$ (Algo.~\ref{alg:collaborative_exploration}, lines 7–8). The proposed deployment mechanism is triggered when the expected exploration gain in 3D significantly exceeds that in 2.5D, as defined below:} 
    \begin{equation}
        \mathcal{H}(\phi_{L,\textrm{best}}^g, \bar{\phi}_{L,\textrm{best}}^a) =
        \begin{cases} 
            1, & \text{if } \phi_{L,\textrm{best}}^g \leq e^{-\gamma_D}\bar{\phi}_{L,\textrm{best}}^a, \\
            0, & \textrm{otherwise},
            \label{eq:deployment_trigger_mechanism}
        \end{cases}
    \end{equation}
    where a value of $1$ indicates that the mechanism is triggered and $\gamma_D>0$ controls the deployment penalty (Algo.~\ref{alg:collaborative_exploration},~lines~9–10). 
    \addition{In essence, the aerial robot is deployed when the potential exploration gain in 3D space exceeds that of the ground robot—even if the ground robot still has viable exploration options. This typically occurs in environments where the aerial robot can more efficiently explore complex 3D structures such as steep slopes, narrow passages, or large vertical spaces beyond the ground robot’s effective reach.}

    Integrated with GBPlanner, the proposed compression method and keyframing strategy enable efficient map-sharing during and after deployment. Each robot $i \in \{g,a\}$ uses a VAE $\{\mathcal{E}_i,\mathcal{D}_i\}$ to encode its range image $\mathbf{x}_i$, derived from point cloud $\mathcal{P}_0^i$ projection, into a latent vector $\mathbf{z}_i$. Combined with the sensor pose $\xi_0^i$ at the time of capture, this forms a keyframe $k_i = \{\mathbf{z}_i,\mathbf{\xi}_0^i\}$. A new keyframe is added to the set $\mathbb{K}_i$ whenever the robot's translation or rotation exceeds predefined thresholds $\tau_t$ or $\tau_r$ relative to the last keyframe pose $\xi_k^i$ \mbox{(Algo.~\ref{alg:keyframing})}. The ground robot maintains its keyframe set $\mathbb{K}_g$ \mbox{(Algo.~\ref{alg:collaborative_exploration}, line 11)}, and shares a subset to initialize the aerial robot’s volumetric map upon deployment.

    \textbf{Deployment Phase:} During the aerial robot's deployment, a co-localization technique \textemdash triggered only once at the deployment time \textemdash enables both robots to operate within a shared inertial frame $\mathcal{I}$. This allows each robot to independently update its map by incorporating both local observations and shared data during post-deployment, while leveraging the collaborative map-sharing solution. \addition{Consistent timestamping across platforms—enabled by Chrony—ensures accurate and reliable map fusion.} Building on~\cite{khattak2020complementary}, the ground system shares: a) a dense local point cloud map $\mathcal{M}$, and b) its current pose $\xi_0^g$, which serves as an initial estimate for the aerial robot's localization. This enables co-localization by iteratively aligning the aerial robot's scan to $\mathcal{M}$ through point-to-line and point-to-plane minimization. Once co-localization is achieved, the aerial robot shares back the transform $T_a^g$ between the two LiDAR frames \mbox{(Algo.~\ref{alg:collaborative_exploration}, line 12)}.

    After co-localization, the ground robot sends the latest $N_k$ keyframes to the aerial robot, which uses the ground robot's decoder $\mathcal{D}_g$ to reconstruct the corresponding point clouds. These, along with their poses, are integrated into the aerial robot's volumetric map to enable collaborative planning (Algo.~\ref{alg:collaborative_exploration}, line 13). The ground robot also shares the target point $\mathbf{p}_{\textrm{target}}^a$, guiding the aerial platform to its initial position. Using its local graph $\mathbb{G}_L^a$, the aerial robot then computes a path to the target and initiates exploration (Algo.~\ref{alg:collaborative_exploration},~line~14). \addition{To prevent mapping artifacts when both robots operate in overlapping areas, Voxblox’s TSDF integration filters out transient objects such as the other robot by repeatedly updating free space, ensuring only persistent structures remain in the map.}

    \textbf{Post-Deployment Phase:} 
    \addition{After the ground robot shares the necessary data and the aerial robot reaches its target, both begin independent exploration. During this phase, they exchange keyframes when within communication range $r_c$ and store them for later transfer when out of range. To address endurance constraints, an energy-aware regrouping strategy ensures both robots return to a common point before battery depletion. A tunable time budget $t_b$, set below the battery life of the robot with the least remaining capacity, guarantees a timely return. Initially, the regrouping point is set to the deployment location. When in communication range, the ground robot estimates travel times $\mathbb{T}_g$ to its keyframes using shortest paths from its global graph $\mathbb{G}_G^g$, assuming constant velocity, and shares them with the aerial robot (Algo.~\ref{alg:collaborative_exploration}, lines 25–26). The aerial robot then computes its own times $\mathbb{T}_a$ to reachable keyframes—those with collision-free paths in $\mathbb{G}_G^a$—and selects a new regrouping point $\mathbb{K}_{g,\text{best}}$ according to (Algo.~\ref{alg:collaborative_exploration}, lines 30-32):    
    \begin{equation}
        \mathbb{K}_{g,\text{best}} = \mathbb{K}_{g,\kappa} \text{ where } \kappa=\operatorname*{arg\,min}_{q\in\{1,\ldots,|\mathbb{K}_g|\}} \big(\max\{\mathbb{T}_{g,q},\mathbb{T}_{a,q}\}\big)
    \end{equation}
    where $|\mathbb{K}_g|$ is the number of ground robot keyframes. The aerial robot shares the selected regrouping point with the ground robot (Algo.~\ref{alg:collaborative_exploration}, lines 27, 33). At each planning step, both robots estimate the time needed to complete their next path and return. If this total exceeds the time budget, regrouping is triggered. The mission concludes once both robots return to the regrouping point and exchange all remaining data (Algo.~\ref{alg:collaborative_exploration}, lines 34).}    
    
    \section{Evaluation Studies}   
\label{sec:evaluation_studies}
    A marsupial ground-aerial robot team was employed to evaluate the proposed approach in both simulation and real-world experiments. The team consists of a legged ground robot, \mbox{ANYmal-D}, and a aerial robot, RMF-Owl~\cite{petris2022rmf}, operating in a marsupial configuration (Fig.~\ref{fig:intro}). \mbox{ANYmal-D} measures \mbox{$0.93$ m $\times$ $0.53$ m $\times $ $0.80$ m} (L $\times$ W $\times$ H), is equipped with a Velodyne VLP-16 LiDAR (FoV: [$360^{\circ}$, $30^{\circ}$], range: $100$~m), and serves as a carrier platform. It runs on $2\times$ 8th Gen Intel Core™ i7 CPUs. RMF-Owl features a collision-tolerant frame (\mbox{$0.38$ m $\times$ $0.38$ m $\times$ $0.24$ m}) and an Ouster OS0-64 LiDAR (FoV: [$360^{\circ}$, $90^{\circ}$], range: $50$ m), powered by a Khadas VIM4 with $4\times$ 2.2GHz Cortex-A73 and $4\times$ 2.0GHz Cortex-A53 cores. It is mounted on the ground robot via a dedicated marsupial mechanism. \addition{Each robot is pre-equipped with its own trained VAE encoder for data compression and uses the other robot’s decoder for decompression.} Communication is handled over WiFi using the NimbRo framework. All processes—co-localization, map sharing, target sharing, and collaborative exploration—run fully onboard and in real time.

\subsection{Training Methodology}
   To account for differences in LiDAR characteristics, separate VAE models were trained on platform-specific datasets composed of simulated and real-world range images from diverse environments, including caves, confined spaces, and complex buildings. The aerial dataset included $\sim36,000$ images ($\sim26,000$ simulated), while the ground dataset had $\sim25,000$ ($\sim21,000$ real). Each model was trained independently for $20$ epochs using the Adam optimizer (learning rate $10^{-4}$, batch size $16$), with a $90\%-10\%$ train-test split.

\subsection{Ablation Study}
    \begin{figure}[!t]
        \centering
        \includegraphics[clip, trim=0cm 0.15cm 0cm 0cm, width=1\linewidth]{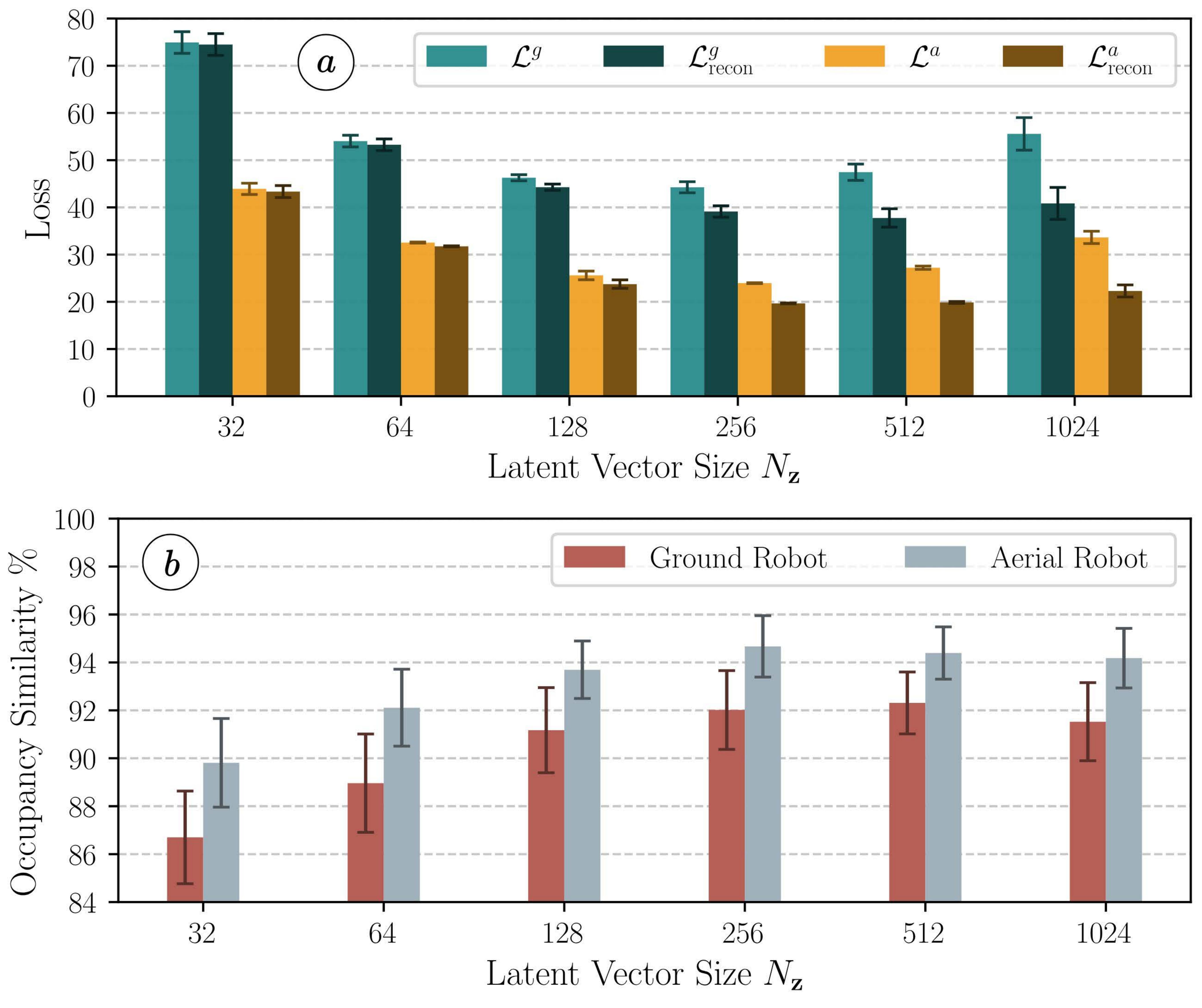}
        \vspace{-2em}
        \caption{\addition{Comparison of (a) overall and reconstruction losses ($\pazocal{L}^g$, $\pazocal{L}_{\textrm{recon}}^g$ for ground; $\pazocal{L}^a$, $\pazocal{L}_{\textrm{recon}}^a$ for aerial), and (b) occupancy similarity for the ground $\{\mathcal{E}_g, \mathcal{D}_g\}$ and aerial $\{\mathcal{E}_a, \mathcal{D}_a\}$ VAE models. Metrics are computed on the test set across latent sizes  ($N_{\textbf{z}}$) and voxel sizes ($s_{vxl}$). Each bar shows the average over three voxel sizes.}}
        \label{fig:ablation_study}
        \vspace{-1ex}
    \end{figure}

    \addition{An ablation study was conducted to evaluate the effects of latent dimensionality and voxel resolution on volumetric map reconstruction. VAE models were trained with latent sizes $N_{\textbf{z}} \in \{32, 64, 128, 256, 512, 1024\}$ and voxel sizes $s_{\text{vxl}}~\in~\{0.2, 0.3, 0.4\}$m for both robots. As shown in Figure~\ref{fig:ablation_study}a, the overall loss decreased with larger latent sizes, reaching a minimum at $256$, then increased due to rapidly growing KL divergence, which led to over-regularization and degraded reconstruction quality \cite{higgins2017betavae}.}

    \addition{To assess how well compressed data retains task-relevant spatial information, we introduce an occupancy similarity metric, which compares voxel-wise occupancy between maps generated from original and reconstructed range images using a k-nearest voxel approach. This metric reflects planning utility more directly than image-space fidelity. As shown in Figure~\ref{fig:ablation_study}b, occupancy similarity trends align with reconstruction loss, peaking at latent sizes of $256$ (aerial) and $512$ (ground), indicating that $256$ is the smallest latent size offering a strong trade-off between spatial consistency and bandwidth. This supports our choice to remap data into a voxel-aware format and evaluate models using both image and task-specific metrics. The approach achieves high compression rates—$337\!\!:\!\!1$ (ground) and $384\!\!:\!\!1$ (aerial)—enabling low-bandwidth communication (e.g., LoRa) for real-time multi-robot coordination in challenging environments.}

\subsection{Simulation Studies}
    
    \begin{table}[]
        \centering
        \begin{threeparttable}
        \caption{\small Parameters for Both Experiments}
        \label{tab:parameters}
        \renewcommand{\arraystretch}{1.1} 
            \begin{tabular}{lcc}
                \toprule
                \textbf{Parameter} & \textbf{Ground Robot} & \textbf{Aerial Robot} \\ \hline
                Size of range image $\mathbf{x}$ ($H\times W$)          & $16\times1800$        & $64\times512$         \\ 
                Voxel size $s_{vxl}$                                    & $0.2$ cm              & $0.2$ cm              \\ 
                Maximum range of image $r_{\text{img}}^{\max}$    & $20$ m                & $20$ m                \\
                Latent space size $N_{\textbf{z}}$                      & $256$                 & $256$                 \\ 
                Deployment sensitivity $\gamma_D$                       & $3.5$ / $4.5$         & -                     \\ 
                Translation threshold $\tau_t$                          & $2.0$ m               & $3.0$ m               \\
                Rotation threshold $\tau_r$                             & $0.785$ rad           & $0.785$ rad           \\
                Number of keyframes $N_k\subseteq\mathbb{K}_g$          & $300$ / $10$          & -                     \\
                Communication range $r_c$                               & $50$ m / $10$ m       & $50$ m / $10$ m       \\
                Time budget $t_b$                                       & $2000$ s / $300$ s     & $2000$ s / $300$ s     \\ 
                Nominal speed                                           & $0.7$ m/s             & $1$ m/s               \\ 
                \bottomrule
            \end{tabular}
            \begin{tablenotes}
              \scriptsize
              \item[*] Parameters of simulation and real-world experiments, represented as $\mathfrak{a}$ / $\mathfrak{b}$ for distinct values, or as $\mathfrak{c}$ for common values.
            \end{tablenotes}
        \end{threeparttable}
    \end{table}

    \begin{figure*}[!ht]
        \centering
        \includegraphics[clip, trim=0cm 0cm 0cm 0cm, width=1\linewidth]{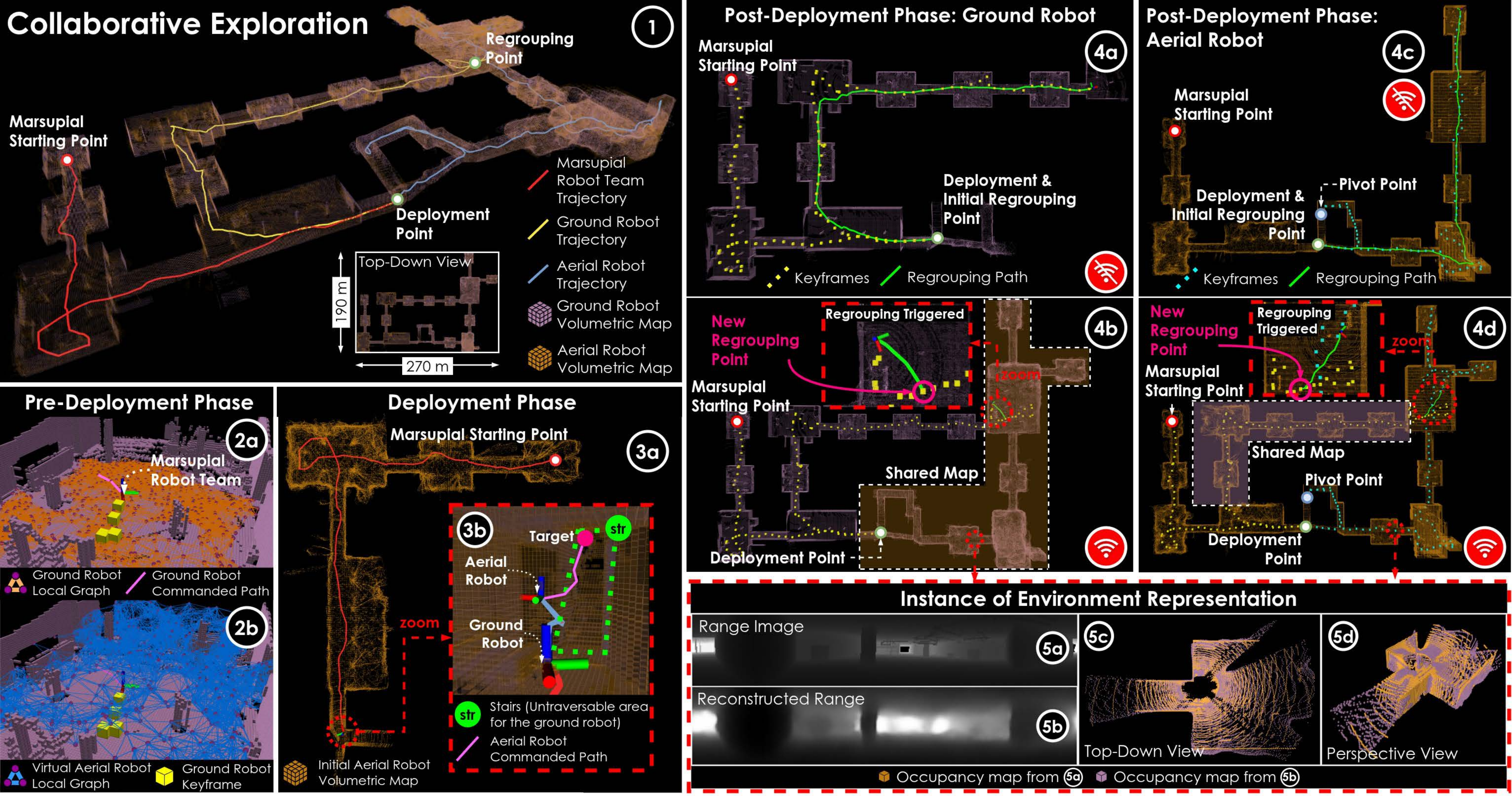}
        \vspace{-2em}
        \caption{\addition{Simulation results showcasing collaborative exploration by a marsupial ground–aerial robot team in a multi-level unknown environment (1). The process includes: (a) Pre-Deployment—the ground robot explores while carrying the aerial robot, builds dense graphs, evaluates potential aerial deployment locations, and stores keyframes (2a-2b); (b) Deployment—the ground robot shares the map and assigns a target to the aerial robot (3a-3b); and (c) Post-Deployment—both robots explore independently, share keyframes, and select a new regrouping point when within communication range (4a-4d). Finally, an instance of the aerial robot's range image (5a), its reconstruction by the ground robot (5b), and the corresponding occupancy maps (5c–5d) are shown.}}
        \label{fig:simulation_results}
        \vspace{-3ex}
    \end{figure*}

    \addition{We validated our approach using a large-scale Gazebo simulation in a multi-level building with interconnected rooms, hallways, stairs, and ramps. The marsupial ground–aerial team coordinated successfully, meeting at an intermediate location to reduce return times and improve energy efficiency via the proposed opportunistic regrouping strategy. Robot behavior and sensors were accurately modeled using the ANYmal and RotorS simulators (see Sec.~\ref{sec:evaluation_studies}). Figure~\ref{fig:simulation_results} shows the collaboratively explored volumetric maps, including key elements such as local and virtual graphs, deployment/regrouping points, shared maps, keyframes, and onboard representations. Co-localization was unnecessary as both robots operated in a shared reference frame provided by the simulation environment. The effectiveness of data sharing is highlighted by pivot points where the aerial robot redirected to unexplored areas after detecting overlap with the ground robot's map. The visible differences between the original and reconstructed range images (Fig.~5, 5a–5b) stem from the task-driven VAE's high compression ratio, which prioritizes occupancy-relevant features over raw geometric detail. Table~\ref{tab:parameters} lists the mission parameters used during the 42-minute operation, which included 12 minutes in marsupial mode and 30 minutes of independent exploration per robot. Table~\ref{tab:quantitative_comparison} presents how compression and keyframing techniques contributed to bandwidth reduction. Latent encoding significantly lowers data rates, and keyframing ensures only essential updates are shared, minimizing communication overhead for bandwidth-constrained scenarios.}

\subsection{Experimental Results}
\begin{figure*}[!ht]
        \centering
        \includegraphics[clip, trim=0cm 0cm 0cm 0cm, width=1\linewidth]{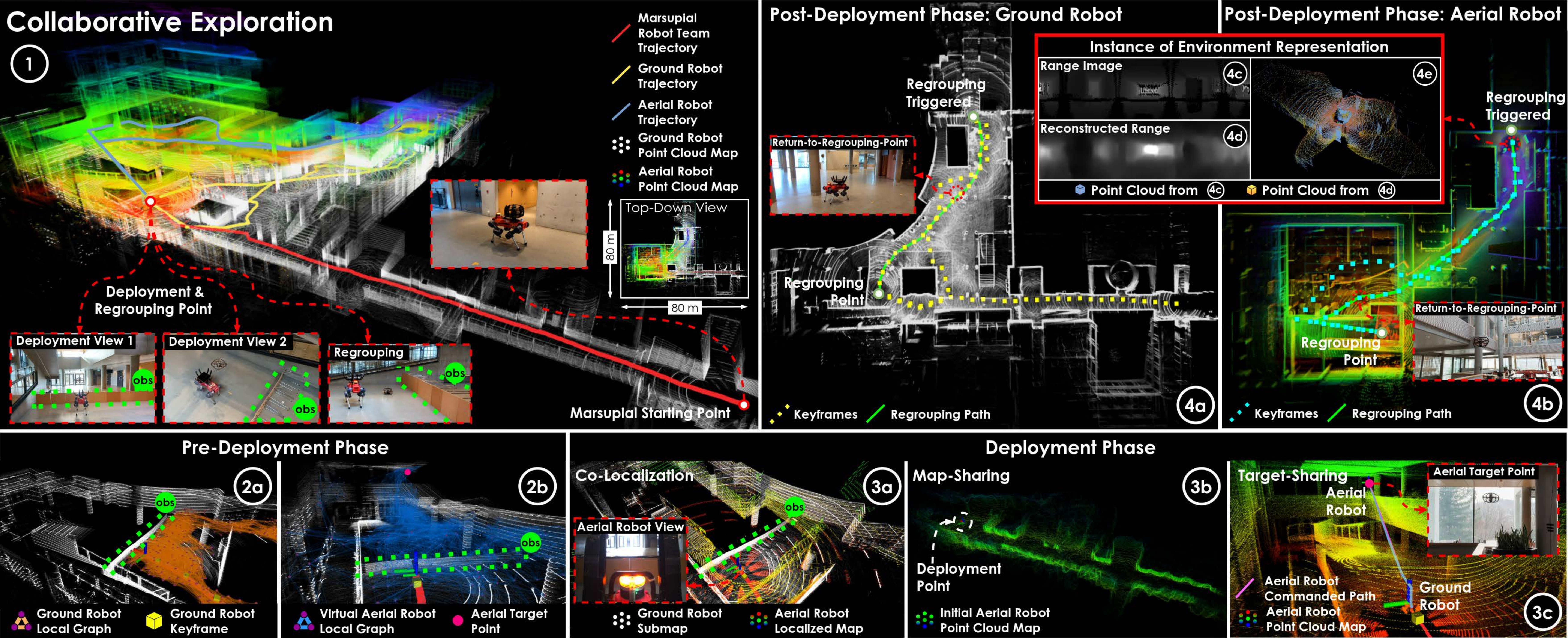}
        \vspace{-2em}
        \caption{Experimental results from the collaborative exploration of the marsupial ground-aerial robot team in a multi-storey university building. The combined map, explored by both robots, is shown in (1) and highlights the start, deployment, and regrouping points. The ground and virtual aerial planning graphs, generated during the pre-deployment phase and used to trigger deployment, are depicted in (2a)–(2b). The processes of co-localization during the aerial robot deployment (3a), map-sharing (3b), and target-sharing (3c) are also presented. The individual exploration maps along with regrouping path are illustrated in (4a)–(4b). An instance of the aerial robot’s range image (4c), the associated reconstructed image on the ground robot (4d), and the corresponding point clouds (4e) are provided.}
        \label{fig:experimental_results}
        \vspace{-3ex}
    \end{figure*}

    To demonstrate the applicability of the proposed method, we conducted real-world experiments in a multi-storey university building using a marsupial ground–aerial robot team. The ground robot was equipped with a custom 3D-printed deployment mechanism (Fig.~\ref{fig:intro}), featuring a flat platform, motorized brackets powered by ANYmal, and a secure mounting system for the aerial robot. The mission began at a designated start point, with the team exploring until the ground robot encountered an untraversable area due to a physical obstacle (marked “obs” in Fig.~\ref{fig:experimental_results}), triggering aerial deployment. During deployment, co-localization, map sharing, and target sharing were performed. \addition{The aerial robot processed each received keyframe in $100$ ms.} After deployment, both robots explored independently, exchanging data when within network range. Upon regrouping, they returned to the deployment point as they were out of range and completed the mission by merging their explored maps. \addition{The differences between original and reconstructed range images (Fig.~\ref{fig:experimental_results}, 4c–4d) reflect the aggressive task-driven compression, which retains mapping-relevant structure over visual detail. Portions of the drone cage visible in the aerial images were masked and inpainted prior to encoding to avoid corrupting the latent space.} The mission lasted $10$ minutes, during which the ground and aerial robots explored approximately $\sim6,600$ $\text{m}^3$ and $\sim7,500$ $\text{m}^3$, respectively. Table~\ref{tab:parameters} summarizes mission settings. \addition{For each keyframe, generation took $10$ ms on the ground robot and $400$ ms on the aerial robot, with data rates of $0.187 $ kB/s and $0.276$ kB/s, respectively.} Figure~\ref{fig:experimental_results} illustrates key stages of the collaborative exploration process.
    
    \addition{Simulation and real-world results show that the proposed framework scales well in distance and application scope. The energy-aware regrouping enables flexible coordination without returning to the deployment point if it is possible, supporting longer missions. Its modular design and bandwidth-efficient map sharing make it suitable for scenarios like subterranean exploration, industrial inspection, and disaster response in GPS- or communication-limited environments.}

    \begin{table}[!t]
        \caption{\small Quantitative Comparison of Data Transmission Rates}
        \label{tab:quantitative_comparison}
        \resizebox{\linewidth}{!}{
            \begin{tabular}{lr}
               \toprule
                \textbf{Transmission Mode}                          & \textbf{Data Rates (kB/s)}    \\ \midrule
                Raw point cloud transmission (10 Hz)                & $3375$                        \\
                Keyframed raw point cloud transmission              & $58.387 $                     \\
                Latent vector transmission (10 Hz)                  & $10$                          \\
                \textbf{Keyframed latent space transmission}        & $\mathbf{0.173}$              \\
                \bottomrule
            \end{tabular}
        }
    \end{table}

    \section{Conclusion}
    \label{sec:conclusion}
    In this paper, we presented a collaborative exploration approach for a marsupial ground-aerial robot team. Through a bandwidth-efficient, task-driven map-sharing solution, both robots can plan based on not only their local observations but also on shared information, enabling more efficient exploration. An ablation study further highlights the trade-off between the size of shared data and the quality of the reconstructed environment. Both simulation and real-world experiments were conducted to validate the proposed approach.

    \bibliographystyle{IEEEtran}
    \bibliography{references}
    
\end{document}